\documentclass[letterpaper, 10 pt, conference]{ieeeconf}  

\IEEEoverridecommandlockouts                              

\usepackage{amsmath,amssymb,amsfonts}
\usepackage{algorithmic}
\usepackage{graphicx}
\usepackage{textcomp}
\usepackage{xcolor}
\usepackage{lipsum}

\usepackage[style=ieee]{biblatex}

\DeclareSourcemap{
  \maps{
    \map{
      \pertype{article}
      \step[fieldset=language, null]
      \step[fieldset=url, null]
      \step[fieldset=doi, null]
      \step[fieldset=issn, null]
      \step[fieldset=isbn, null]
      \step[fieldset=note, null]
      \step[fieldset=editor, null]
      \step[fieldset=urldate, null]
      \step[fieldset=file, null]
    }
  }
}
\DeclareSourcemap{
  \maps{
    \map{
      \pertype{inproceedings}
      \step[fieldset=language, null]
      \step[fieldset=url, null]
      \step[fieldset=doi, null]
      \step[fieldset=issn, null]
      \step[fieldset=isbn, null]
      \step[fieldset=note, null]
      \step[fieldset=editor, null]
      \step[fieldset=urldate, null]
      \step[fieldset=file, null]
    }
  }
}
\DeclareSourcemap{
  \maps{
    \map{
      \pertype{incollection}
      \step[fieldset=language, null]
      \step[fieldset=url, null]
      \step[fieldset=doi, null]
      \step[fieldset=issn, null]
      \step[fieldset=isbn, null]
      \step[fieldset=note, null]
      \step[fieldset=editor, null]
      \step[fieldset=urldate, null]
      \step[fieldset=file, null]
    }
  }
}

\title{\LARGE \bf
How Strong a Kick Should be to Topple \\Northeastern's Tumbling Robot?
}

\author{Adarsh Salagame$^1$, Neha Bhattachan$^{1}$, Andre Caetano$^{1}$, Ian McCarthy$^{1}$, Henry Noyes$^{1}$, \\ Brandon Petersen$^{1}$,
Alexander Qiu$^{1}$, Matthew Schroeter$^{1}$, Nolan Smithwick$^{1}$, \\ Konrad Sroka$^{1}$, Jason Widjaja$^{1}$, Yash Bohra$^{1}$, 
Kaushik Venkatesh$^{1}$, Kruthika Gangaraju$^{1}$\\ Paul Ghanem$^{1}$, Ioannis Mandralis$^{2}$, Eric Sihite$^{2}$, Arash Kalantari$^{3}$, and Alireza Ramezani$^{1*}$
\thanks{$^{1}$Authors are with the Silicon Synapse Labs, Department of Electrical and Computer Engineering, Northeastern University, Boston, USA. Emails: 
        {\tt\small salagame.a, a.ramezani@northeastern.edu}}%
\thanks{$^{2}$Author is with the Department of Aerospace Engineering, California Institute of Technology, Pasadena, USA. Email: 
        {\tt\small imandralis, esihite@caltech.edu}}%
\thanks{$^{3}$Author is with Jet Propulsion Laboratory, Pasadena, USA. Email: 
        {\tt\small arash.kalantari@jpl.nasa.gov}}%
\thanks{$^{*}$Corresponding author.}%
}

\begin{document}

\maketitle
\thispagestyle{empty}
\pagestyle{empty}

\begin{abstract}
Rough terrain locomotion has remained one of the most challenging mobility questions. In 2022, NASA's Innovative Advanced Concepts (NIAC) Program invited US academic institutions to participate NASA's Breakthrough, Innovative \& Game-changing (BIG) Idea competition by proposing novel mobility systems that can negotiate extremely rough terrain, lunar bumpy craters. In this competition, Northeastern University won NASA's top Artemis Award award by proposing an articulated robot tumbler called COBRA (Crater Observing Bio-inspired Rolling Articulator). This report briefly explains the underlying principles that made COBRA successful in competing with other concepts ranging from cable-driven to multi-legged designs from six other participating US institutions. 
\end{abstract}


\section{Introduction}

Rough terrain locomotion has remained one of the most challenging mobility questions to date. Many different designs have been proposed so far. Perhaps, the most promising designs constitute legged robots that intermittently interact with their environment to translate the center of mass relative to the ground substrate. Legged systems can walk over bumpy surfaces by exploiting the contact-rich nature of their locomotion as opposed to wheeled systems with fixed contact points. However, legged locomotion over steep slopes with bumpy surface pose different challenges that so far have remained unexplored. 

Bumpy steep slopes are abundant on Earth. However, perhaps, the most famous examples of these environments with significant scientific importance are in outer space. The lunar surface is covered with craters from many years of bombardment by small and large meteors. Some of these craters are a few tens of kilometers in diameter and tens of degrees on their surface slopes. Their surface is exceptionally bumpy and covered with porous and fluffy regolith (lunar soil), which makes locomotion even harder. 

In 2022, NASA's Innovative Advanced Concepts (NIAC) Program invited US academic institutions to participate NASA's Breakthrough, Innovative \& Game-changing (BIG) Idea competition \cite{noauthor_big_nodate} by proposing novel mobility systems that can negotiate the lunar craters. In this competition, Northeastern University won NASA's top award, the Artemis Award, by proposing an articulated robot tumbler called COBRA (Crater Observing Bio-inspired Rolling Articulator). This report briefly explains the underlying principles that made COBRA successful in competing with other concepts ranging from cable-driven to multi-legged designs from six other participating US institutions.      

\begin{figure}
\vspace{0.08in}
    \centering
    \includegraphics[width=1.0\linewidth]{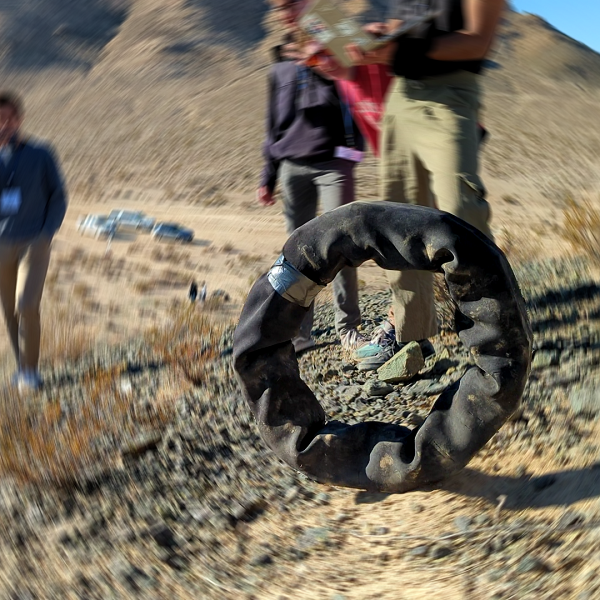}
    \caption{Shows COBRA, Crater Observing Bio-inspired Rolling Articulator, performing tumbling locomotion in NASA's Innovative Advanced Concept Competitions in Pasadena, California, in November 2022.}
    \label{fig:cover}
\vspace{-0.08in}
\end{figure}


Tumbling robots are not new. The robotic community has endorsed the merits and limitations of these systems. Passive tumbling systems such as the NASA/JPL Mars Tumbleweed Rover \cite{behar_nasajpl_2004} use very little energy for locomotion, making them attractive choices for remote exploration where energy efficiency is crucial. Active rolling spherical robots such as MIT's Kickbot \cite{batten_kickbot_nodate} have a low center of gravity and can move in any direction, making them robust to external perturbations and uneven terrain. The ability of spherical robots to roll in any direction also gives them excellent maneuverability in tight spaces. 

Tumbling, however, comes with its challenges. Passive rolling robots typically trade controllability for energy efficiency, relying instead on uncontrollable morphology (or posture) for maneuvering. Also, rolling robots usually use their entire body to execute locomotion. Therefore, the lack of a stable platform for sensors such as cameras makes localization and perception a significant ordeal.

Early robots such as Rollo \cite{halme_motion_1996} and Spherical Mobile Robot (SMR) \cite{reina_rough-terrain_2004} had a spherical shell with a diametric spring-mass system that spun a driving wheel to shift the axis of the mass, creating a mass imbalance that generated movement. These designs were unreliable due to challenges in keeping the driving wheel constantly in contact with the sphere. In addition, a large weight for the central structure was required to generate any meaningful inertia to react with the torque from the driving wheel and propel the system forward.

Robotic systems with similar concepts but replacing the diametric driving wheel, with a car-driven system inside spherical structures were introduced too. Notable examples are the University of Pisa's Sphericle \cite{bicchi_introducing_1997} and Spider Rolling Robot (SRR) from Festo \cite{western_golden_2023}. Sphericle relied on gravity to keep the car wheels in contact with the inside of the sphere and large perturbations led by mobility on rough terrain could dislodge the car and incapacitate the robot. SRR has the car modeled as a 1-dof pendulum on a fixed axis inside the spherical shell, allowing locomotion only in one plane and relying on the inertia of the pendulum to move the robot.  Another means of tumbling locomotion is the shifting of weights inside a rigid spherical shell to precisely control the 3D position of the center of mass. Examples of systems using this concept are the University of Michigan's Spherobot \cite{mukherjee_simple_1999} and the University of Tehran's August Robot \cite{javadi_a_introducing_2004}. With the added dead weights required to control the center of mass, these robots are not the most energy-optimal solution.

A more energy-efficient means of positioning the center of gravity for tumbling is by using deformable structures. Successful examples \cite{tian_dynamic_2015, wei_design_2019,wang_trajectory_2018, wang_dynamics_2018,sastra_dynamic_2009} can be identified that have attempted rolling by articulated structural designs that allow such deformation. Notable examples are Ritsumeikan University's Deformable Robot \cite{sugiyama_crawling_2006}, and Ourobot \cite{paskarbeit_ourobotsensorized_2021} with Articulated closed loop. COBRA also utilizes the concept of deformation of structure to initiate and control tumbling. However, what differentiates COBRA from these examples are: 1) Multi-modal locomotion abilities (sidewinding, slithering, and tumbling), 2) Field-tested capabilities for fast and dynamic tumbling locomotion on bumpy surfaces, 3) Head-tail locking mechanism to form rugged structures for tumbling, and 4) The ability to control posture for active steering in 2 dimensions. Of these, this paper remains focused on tumbling locomotion and provides an overview of the head-tail locking concept.       








This work is organized as follows. First, we provide a brief overview of COBRA's hardware, and in particular, the head-tail locking mechanism. Second, we mathematically describe what made COBRA successful in the BIG Idea competition. We employ Mixed-Hamiltonian-Lagrangain concepts to relate COBRA's kinetic energy to the forces needed to topple it when tumbling. Then, we provide numerical results based on the mathematical frame explained in the math section. Last, we present the experimental results, lessons, and conclusions from our field tests. 

\section{COBRA's Hardware Overview}
\begin{figure*}
\vspace{0.08in}
    \centering
    \includegraphics[width=0.8\linewidth]{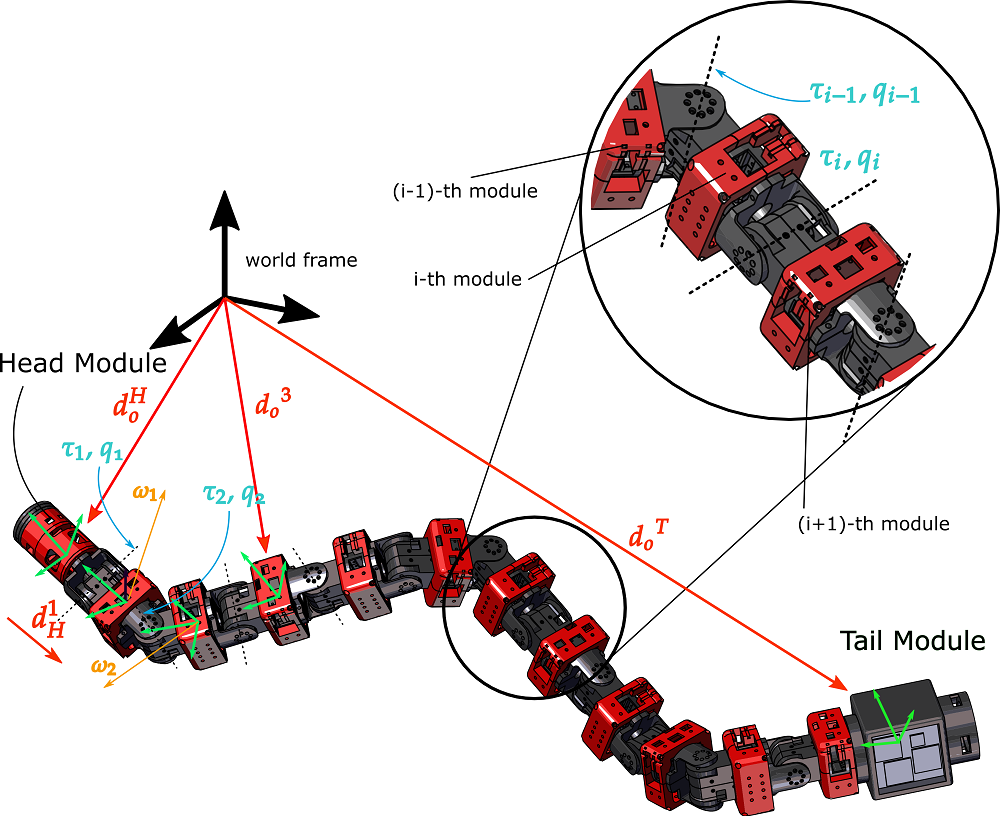}
    \caption{Illustrates the coordinate frames and parameters used for modeling COBRA.}
    \label{fig:cobra-coordinates}
\vspace{-0.08in}
\end{figure*}

As seen in Fig.~\ref{fig:cobra-coordinates}, the COBRA system consists of eleven actuated joints. The front of the robot consists of a head module containing the onboard computing of the system, a radio antenna for communicating with a lunar orbiter, and an inertial measurement unit (IMU) for navigation. At the tail end, there is an interchangeable payload module which, for our mission, contains a neutron spectrometer to detect water ice. The rest of the system consists of identical 1-dof joint modules (Fig.~\ref{fig:cobra-coordinates}) that each contain a joint actuator and a battery.

\subsection{Head-tail Locking Mechanism}

\begin{figure}
\vspace{0.08in}
    \centering
    \includegraphics[width=0.8\linewidth]{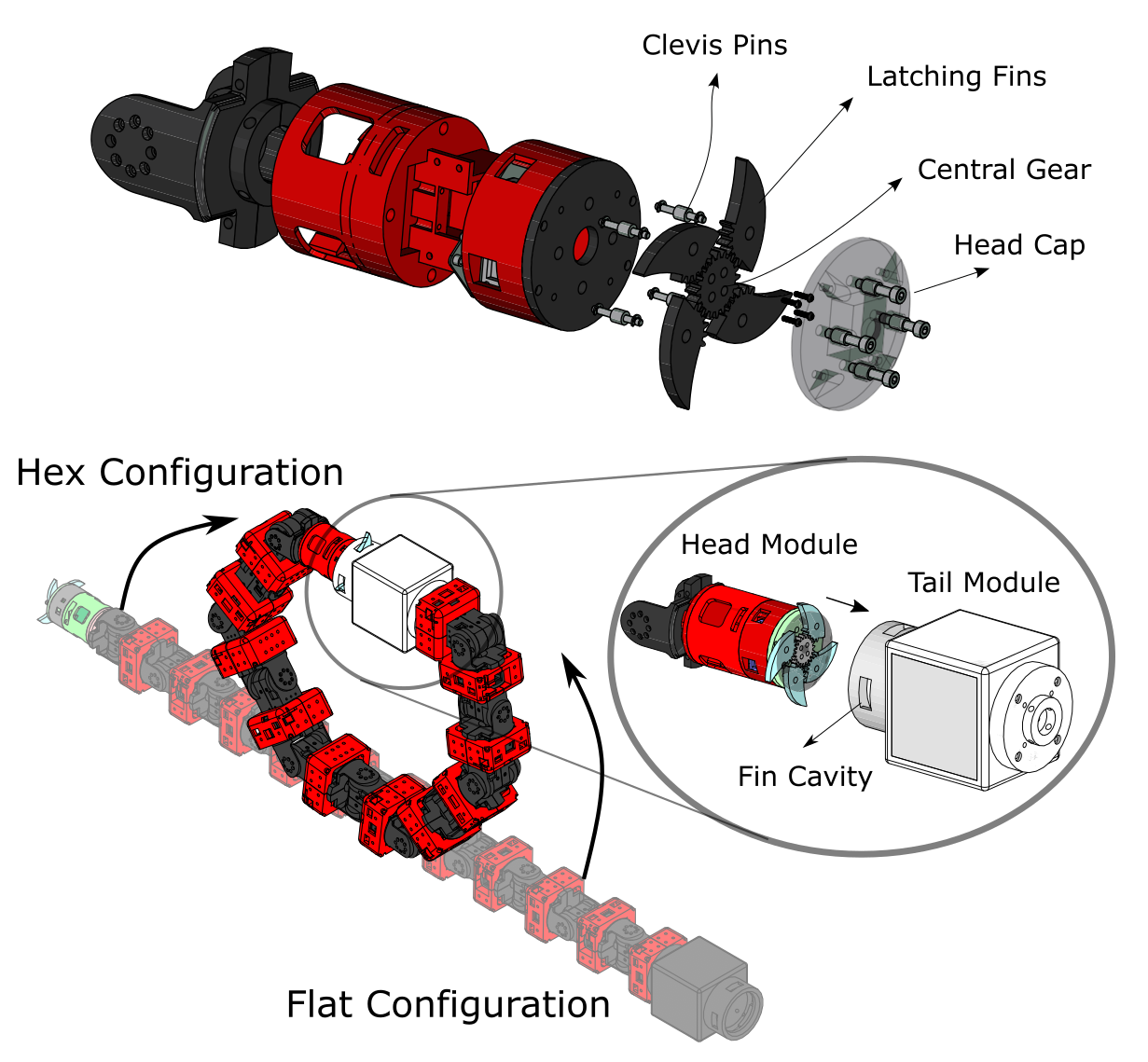}
    \caption{Detailed illustration of the head-tail locking mechanism.}
    \label{fig:head}
\vspace{-0.08in}
\end{figure}

In addition to the eleven identical modules, COBRA features a distinct module at the snake’s head, aptly referred to as the “head module,” and similarly, a “tail module” at the snake’s tail end. The head module is shown in Fig.~\ref{fig:head}. The primary purpose of these unique modules are to connect together to form a loop prior to the onset of  tumbling mode. The head module acts as the male connector and utilizes a latching mechanism to sit concentrically inside the female tail module. 

The latching mechanism consists of a Dynamixel XC330 actuator, which sits within the head module and drives a central gear. This gear interfaces with the partially geared sections of four fin-shaped latching “fins.” The curved outer face of each latching fin has an arc length equal to 1/4 of the circumference of the head module’s circular cross-section. When the mechanism is retracted, these four fins form a thin cylinder that coincides with the head module's cylindrical face. A dome-shaped cap lies on the end of the head module so that the fins sit between it and the main body of the head module. Clevis pins are used to position the fins in this configuration. COBRA’s tail module features a female cavity for the fins. When transitioning to tumbling mode, the head module is positioned concentrically inside the tail module using the joint’s actuators, and the fins unfold into the cavity to lock the head module in place. For the head and tail modules to unlatch, the central gear rotates in the opposite direction, and the fins retract, allowing the system to return to sidewinding mode.

The choice for an active latching mechanism design stemmed from the design requirements and restrictions. Magnets were initially discussed as a passive latching option, however they would not be effective in conjunction with the ferromagnetic regolith. Further, due to the need to stay in a latched configuration even when a large amount of force is applied to the system during tumbling, a passive system was not chosen, for there would be the risk of unlatching during tumbling. 



\section{Modeling}

\begin{figure}
\vspace{0.08in}
    \centering
    \includegraphics[width=1\linewidth]{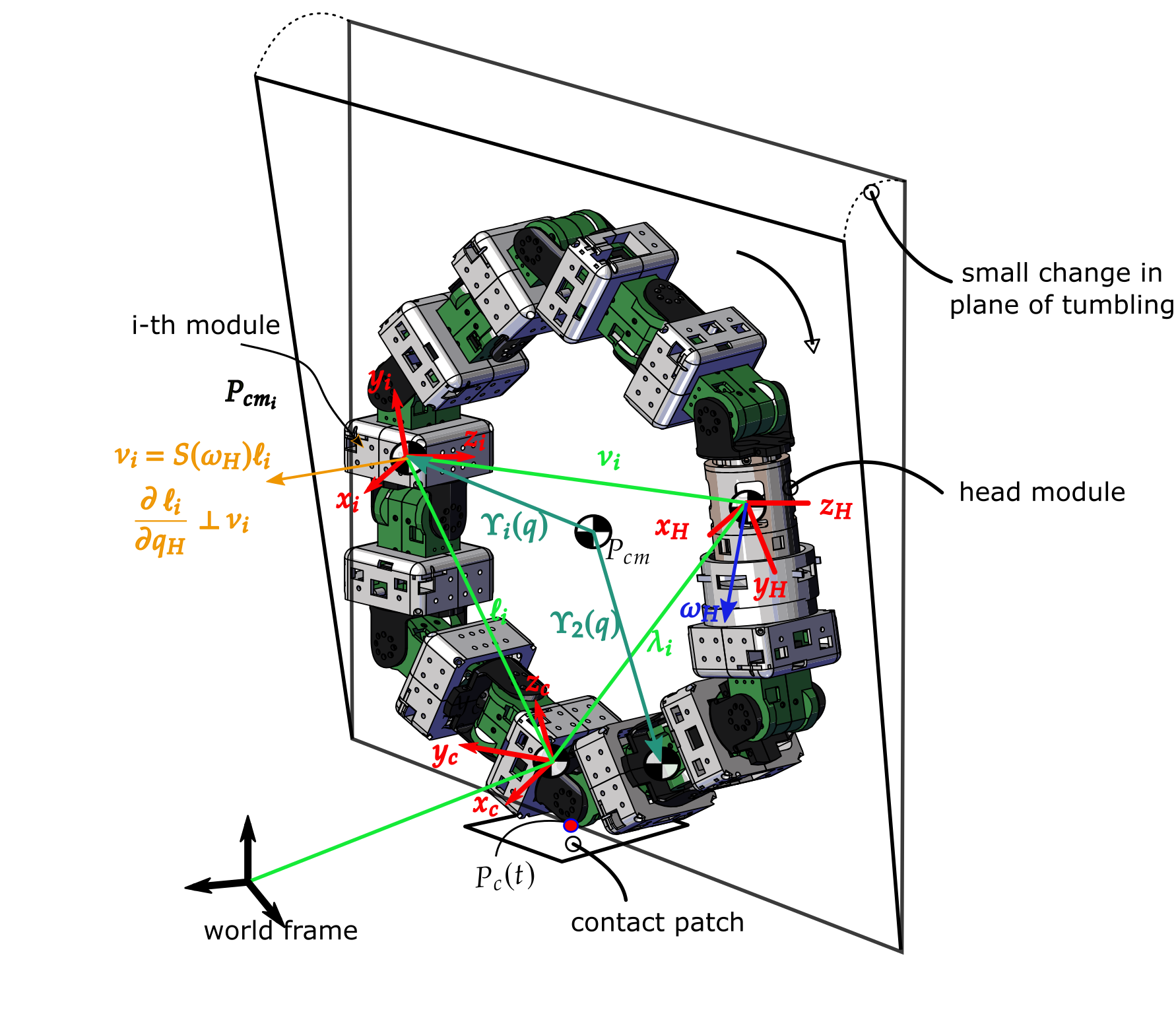}
    \caption{Illustrates $\Upsilon_i$ functions used to simplify kinetic energy}
    \label{fig:tumbling-fbd}
\vspace{-0.08in}
\end{figure}

The aim of this section is to establish a numerical framework to study the tumbling behavior of COBRA. 

\subsection{Kinematics}
Consider the configuration variable vector $q=[\dotsc, q_i, \dotsc, p_H^\top, q_H^\top]^\top$ which embodies the body angles $q_i$, head module position $p_H$, and orientations $q_H$. We use the Euler convention to find the rotation matrix $R_H^0$  
\begin{equation}
R_{H}^{0} (q_H) =R_{z, q_z} R_{y, q_y} R_{x, q_x} 
\label{eq:head-base-transformation}
\end{equation}
\noindent $R_H^0$ gives the representation of the points in the head frame with respect to the world frame. We consider the rotation matrices $R_{i}^{H}$ 
\begin{equation}
R_{i}^{H} =R_{H}^{0}R_{i}^{H},~ i=1,\dots,N
\end{equation}
\noindent $R_i^H$ gives the expression of any points at each modules with respect to the the head frame. Using $R_{i}^{H}$ and $R_{H}^{0}$, we obtain the forward kinematics equation and find the center of mass (CoM) position $p_{cm,i}$ of each module with respect to the world frame
\begin{equation}
p_{cm,i}=R_{i}^{0} p_{cm,i}^{i}+d_{i}^{0}
\end{equation}
\noindent In this equation, $d_{i}^{0}$ denotes the world frame position of the body coordinates attached to i-th module.
Angular velocity of the head module $\omega_H(t)$ and its relationship with the time derivative of the configuration variable $\dot q$ is given by
\begin{equation}
\hat\omega_H(t) = \dot{R}_H^0(t)R_H^0(t),~ \omega_H=\beta_H(q)\dot q
\end{equation}
\noindent $\beta_H(q)$ is the head-module Jacobian matrix. The angular velocity of i-th module and its relation to $\dot q$ are given by
\begin{equation}
\hat\omega_i(t) = \dot{R}_i^0(t)R_i^0(t),~ \omega_i=\beta_i(q)\dot q
\end{equation}
\noindent The world-frame velocity of i-th module CoM $v_{cm,i}$ can be obtained by
\begin{equation}
v_{cm,i}^{0}=\hat\omega_i(t)R_{i}^{0} p_{cm,i}^{i}+\dot{d}_i^0=\sum_{i=1}^N \frac{\partial d_i^0}{\partial q_i} \dot{q}_i
\end{equation}

\subsection{Establishing a Link Between COBRA's Angular Momentum and Total Kinetic Energy}

\begin{figure}
\vspace{0.08in}
    \centering
    \includegraphics[width=1\linewidth]{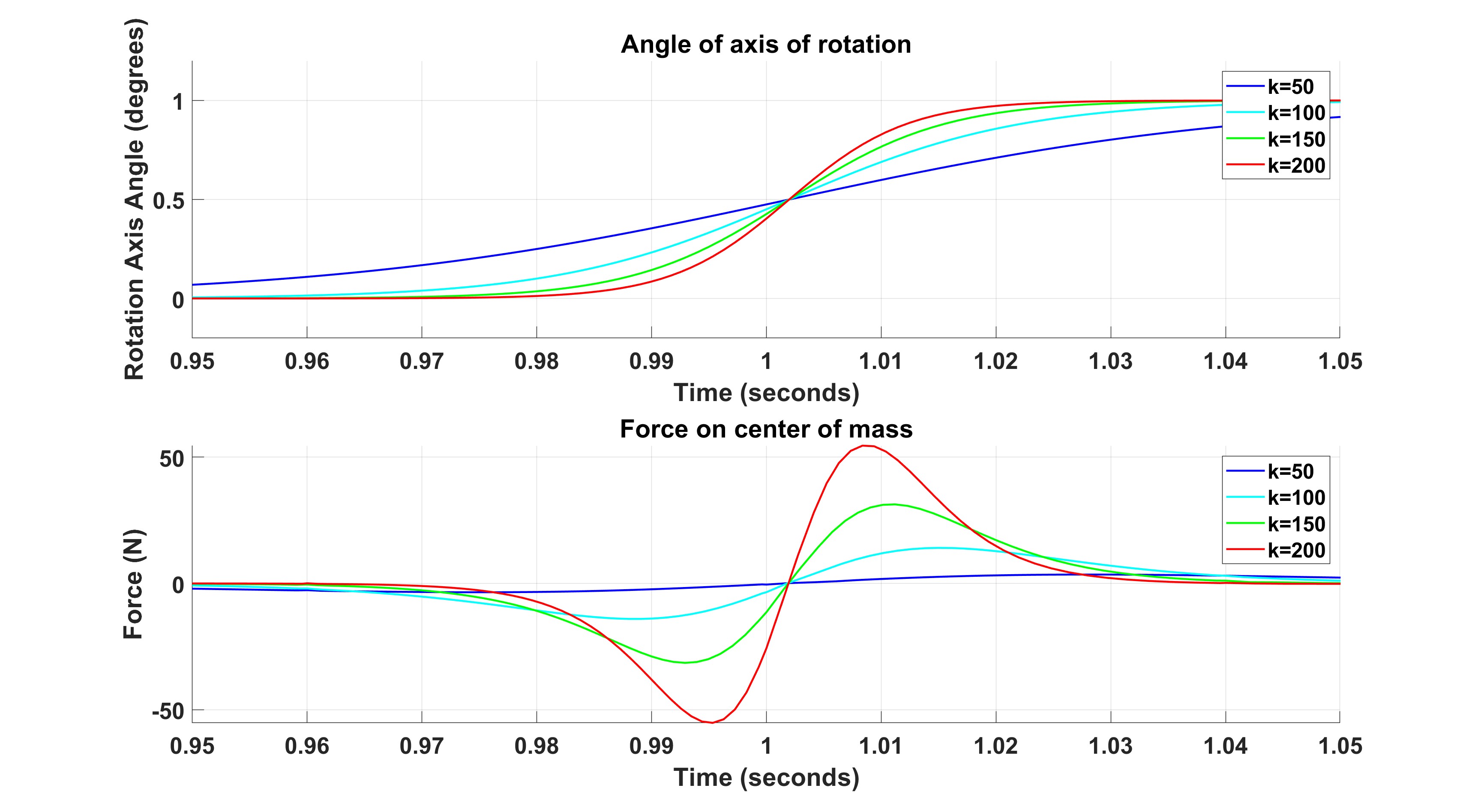}
    \caption{Shows the numerical predictions of the 'Kick' force needed to change one-degree changes in the tumbling articulate structure when rotating at a rate of $10~rad/sec$. 'k' here refers to the growth rate parameter of a standard sigmoid curve with an inflection point of 1 sec. going from zero deg. to 1 deg., used to vary the time taken to change the axis of rotation smoothly.}
    \label{fig:force}
\vspace{-0.08in}
\end{figure}

\begin{figure}
\vspace{0.08in}
    \centering
    \includegraphics[width=1\linewidth]{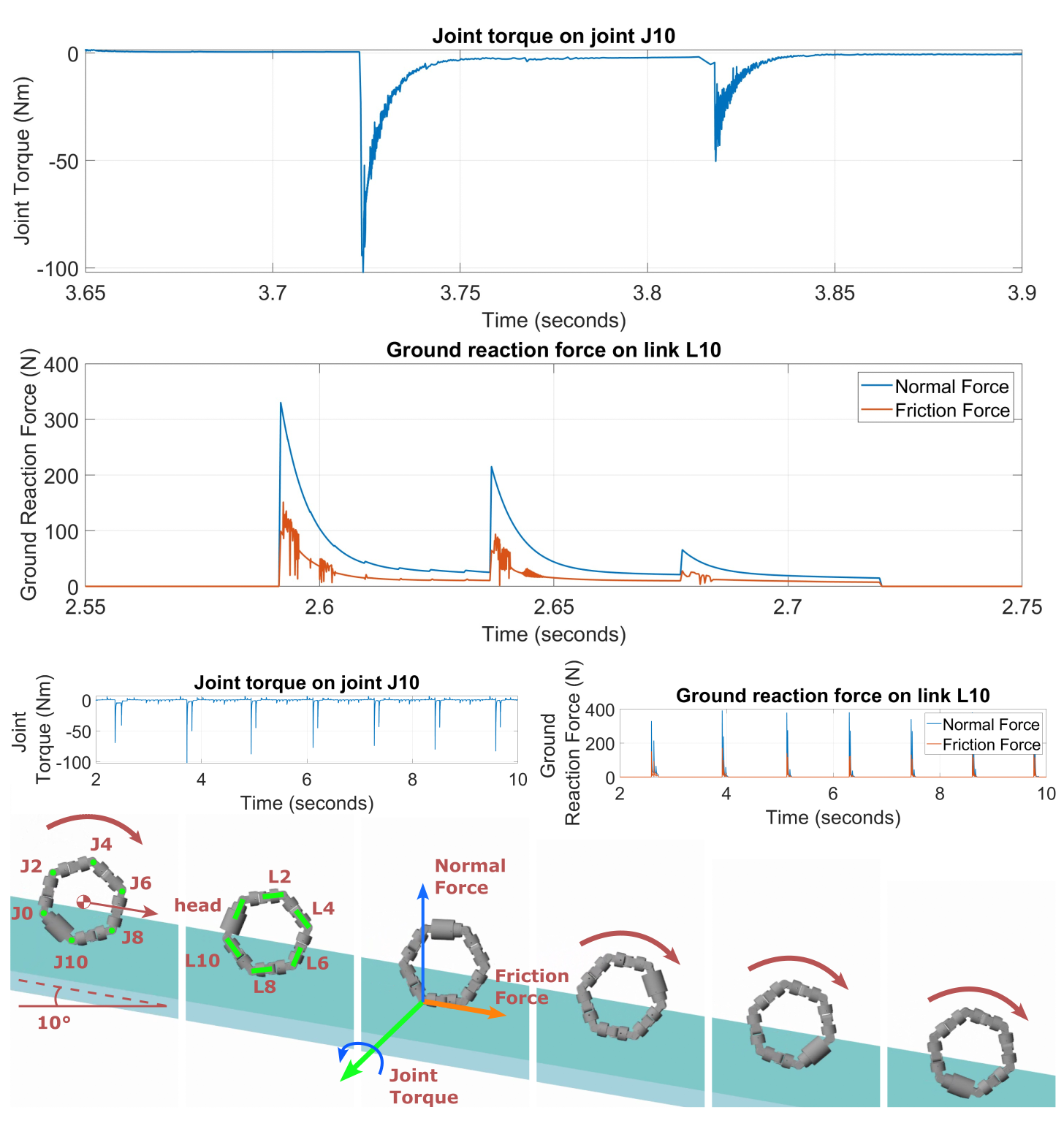}
    \caption{The figure shows the joint torque of a single joint (J10) and ground reaction forces (Normal and Frictional Force Magnitude) on a single link (L10) during one period of contact with the ground while tumbling. The graph below shows the full 10-second simulated period with spikes for each repeated contact.}
    \label{fig:joint-torque-grf}
\vspace{-0.08in}
\end{figure}

Now, consider the angular momentum $\sigma_{a,i}$ about the point $p_a$ on the robot from the i-th module
\begin{equation}
\sigma_{a,i} =m_i\Big(p_{cm,i}-p_a\Big)\textbf{S} \dot p_{cm,i} + J_{cm,i}\omega_i
    \label{eq:ang-momentum}
\end{equation}
\noindent where $m_i$ and $J_{cm,i}$ are i-th module's mass and mass momentum of inertia; $\textbf{S}$ is a skew-symmetric matrix. Using Eq.~\ref{eq:ang-momentum}, the total angular momentum around $p_a$ is given by 
\begin{equation}
\begin{aligned}
\sigma_a =\sum^N_{i=1}\sigma_{a,i}
    =\sum^N_{i=1}\Big(&m_ip_{cm,i}\textbf{S}v_{cm,i} + J_{cm,i}\omega_i\Big) \\&- m_{tot}p_a\textbf{S}v_{cm}
\end{aligned}
    \label{eq:total-ang-momentum-pa}
\end{equation}
\noindent Therefore, the total angular momentum of the robot around its CoM can be obtained by
\begin{equation}
\begin{aligned}
\sigma_{cm}=\sum_{i=1}^{N}\Big(&m_{i} p_{c m, i} \textbf{S} v_{c m, i}+J_{cm, i} \omega_i\Big)\\&-m_{tot} p_{cm} \textbf{S} v_{cm}
\end{aligned}
    \label{eq:total-ang-momentum-com}
\end{equation}
Now, we re-write the forward kinematics equations based on relative position of each module's CoM position with respect to the CoM to find simpler forms of $\sigma_{cm}$. Consider 
\begin{equation}
p_{cm,i} = p_{cm} + \Upsilon_i(q)
\end{equation}
\noindent where $\Upsilon$ is the relative position of i-th module's CoM with respect to the total CoM. Based on the definition of $\Upsilon$, we can write 
\begin{equation}
\begin{aligned}
\sum^N_{i=1}m_i\Upsilon_i(q)=0, & ~~~
\sum^N_{i=1}m_i\dot{\Upsilon_i}(q)=0
\end{aligned}
    \label{eq:upsilon}
\end{equation}
\noindent Now, apply Eq.~\ref{eq:upsilon} to obtain the total kinetic energy $K$ of the tumbling articulated robot
\begin{equation}
\begin{aligned}
K= & \frac{1}{2} \dot{m}_{\text {tot }} \dot{p}_{c m}^{\top} \dot{p}_{c m}+ \frac{1}{2} \dot{q}^{\top} \sum_{i=1}^N\left[m_i \frac{\partial \Upsilon_i^{\top}}{\partial q} \frac{\partial \Upsilon_i}{\partial q}+\beta_i^{\top} J_{cm,i} \beta_i\right] \dot{q}
\end{aligned}
    \label{eq:kinetic}
\end{equation}
\noindent In Eq.~\ref{eq:kinetic}, by applying the properties given by Eq.~\ref{eq:upsilon}, the following term has been canceled out 
\begin{equation}
\sum_{i=1}^N2 m \dot{p}_{cm,i}^\top \frac{\partial \Upsilon_i}{\partial q}\dot{q}=0
\end{equation}
\noindent In addition, using Eq.~\ref{eq:upsilon} and re-writing $\sigma_{cm}$ given by Eq.~\ref{eq:total-ang-momentum-com} yields 
\begin{equation}
\sigma_{cm}=\sum_{i=1}^N \Upsilon_i^{\top} \textbf{S}\left(\omega_H\right) \dot{\Upsilon}_i+ \omega_i^{\top} J_{cm, i}\omega_i 
\end{equation}
\noindent where $\textbf{S}(.)$ is the skew symmetric operator (e.g., turns the vector $\omega_H$ to a skew symmetric matrix). Now, we employ a geometric argument of the cyclic variables $q_H$ (embodies three Euler angles) to obtain the angular momentum $\sigma_{cm}$ of the tumbling articulated structure from its total $K$. 
Consider each module's CoM position with respect to the contact point $p_C$ as shown in Fig.~\ref{fig:tumbling-fbd}
\begin{equation}
    p_{cm,i}=p_C+R_H^0(q_H)\Big(\lambda_i(q_b)-\nu_i(q_b\Big)
\end{equation}
\noindent where $\ell_i=\lambda_i-\nu_i$ is shown in the Fig.~\ref{fig:tumbling-fbd}. The velocity $v_{cm,i}$ based on this new parameterization is given by
\begin{equation}
    v_{cm,i}= R_H^0 \textbf{S}(w_H)\ell_i(q_b)+R_H^0 \dot{\ell}_i(q_b)
\end{equation}
\noindent from the above equation one can see 
\begin{equation}
    \frac{\partial \ell_i}{\partial q_H}= \textbf{S}(w) \ell_i\left(q_b\right)
    \label{eq:partial-ell-partial-cyclic}
\end{equation}
\noindent The geometric interpretation of $\frac{\partial \ell_i}{\partial q_H}$ is shown in Fig.~\ref{fig:tumbling-fbd}. We employ $\frac{\partial \ell_i}{\partial q_H}$ to link $\sigma_{cm}$ to $\frac{\partial K}{\partial \dot q_H}$ which is given by  

\begin{figure*}
\vspace{0.08in}
    \centering
    \includegraphics[width=1\linewidth]{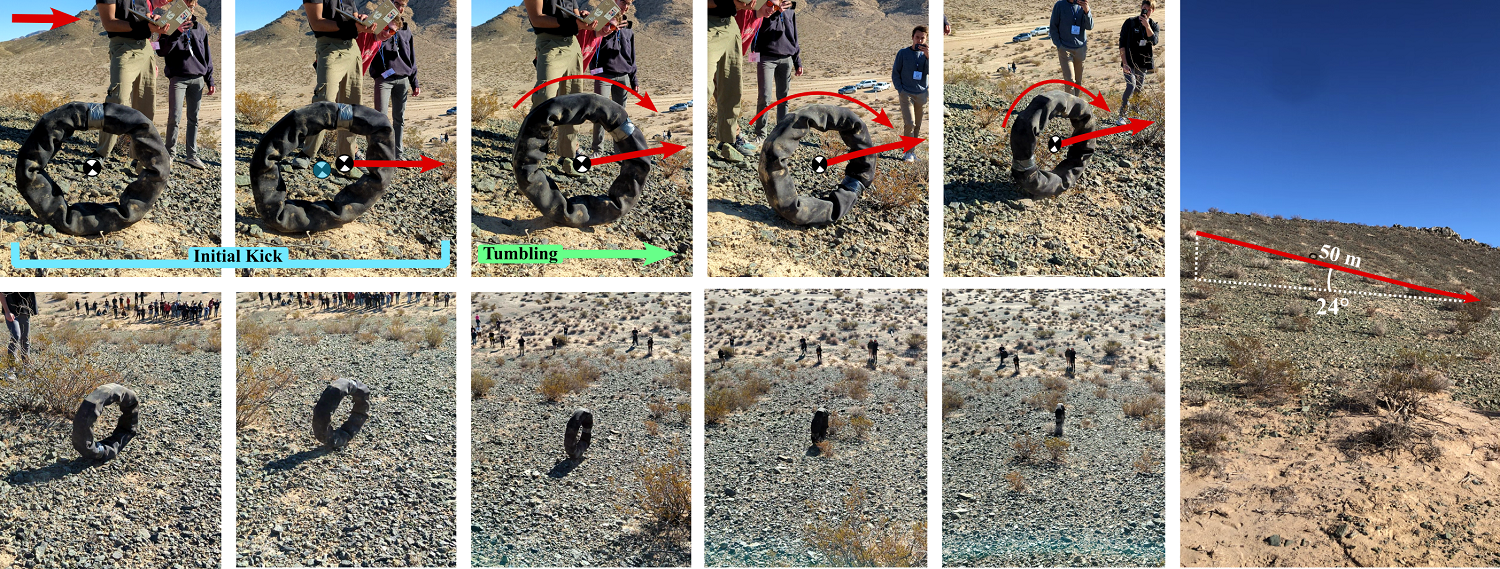}
    \caption{Overlaid snapshots of tumbling}
    \label{fig:exp}
\vspace{-0.08in}
\end{figure*}

\begin{equation}
    \frac{\partial K}{\partial \dot q_H}=\sum_{i=1}^N m_i\left(\frac{\partial \Upsilon_i}{\partial q_H}\right)^\top \dot \Upsilon_i+\left(\frac{\partial \omega_i}{\partial q_H}\right)^{\top} J_{cm,i} \omega_i
    \label{eq:partial-derivative-kinetic}
\end{equation}
\noindent In Eq.~\ref{eq:partial-derivative-kinetic}, by substituting $\frac{\partial \Upsilon_i}{\partial q_H}$ and $\frac{\partial \omega_i}{\partial q_H}$ with Eq.~\ref{eq:partial-ell-partial-cyclic}, $\sigma_{cm}$ given in Eq.~\ref{eq:total-ang-momentum-com} is obtained. This completes our approach to inspect the angular momentum of the tumbling articulated robot based on finding the total kinetic energy.  

Next, we will numerically calculate the force needed to topple the tumbling robot based on obtaining the rates of changes of $\dot \sigma_{cm}(t)$ during tumbling by inspecting the kinetic energy.

\section{Results and Discussion}
The dynamical equations of motion of COBRA when tumbling is given by
\begin{equation}
    \begin{aligned}
        \left[\begin{array}{cc}
D_H & D_{H a} \\
D_{a H} & D_a
\end{array}\right]\left[\begin{array}{l}
\ddot{q}_H \\
\ddot{q}_a
\end{array}\right]&+\left[\begin{array}{l}
H_H \\
H_a
\end{array}\right]=\left[\begin{array}{l}
0 \\
B_a
\end{array}\right]u+\\&
+
\left[\begin{array}{cc}
J_{H}^\top \\ J_{a}^\top
\end{array}\right]F_{GRF}
    \end{aligned}
\label{eq:full-dynamics}
\end{equation}
\noindent where $D_i$, $H_i$, $B_i$, and $J_i$ are partitioned model parameters corresponding to the head 'H' and actuated 'a' joints. $F_{GRF}$ denotes the ground reaction forces. $u$ embodies the joint actuation torques. In this section, we numberically simulate the tumbling dynamics and using the relation between total kinetic energy $K$, inertia matrix $D(q)$, and angular momentum $\sigma_{cm}$, we inspect the force needed to topple COBRA when is tumbling down a steep slope. 

The ground model used in our simulations is given by 
\begin{equation}
\begin{aligned}
    F_{GRF} &= \begin{cases} \, 0 ~~  \mbox{if } p_{C,z} > 0  \\
     [F_{GRF,x},\, F_{GRF,y},\, F_{GRF,z}]^\top ~~ \mbox{else} \end{cases} \\
    F_{GRF,z} &= -k_1 p_{C,z} - k_{2} \dot p_{C,z} \\
    F_{GRF,i} &= - s_{i} F_{GRF,z} \, \mathrm{sgn}(\dot p_{C,i}) - \mu_v \dot p_{C,i} ~~  \mbox{if} ~~i=x, y\\
    s_{i} &= \Big(\mu_c - (\mu_c - \mu_s) \mathrm{exp} \left(-|\dot p_{C,i}|^2/v_s^2  \right) \Big)
\end{aligned}
\end{equation}
\noindent where $p_{C,i},~~i=x,y,z$ are the $x-y-z$ positions of the contact point; $F_{GRF,i},~~i=x,y,z$ are the $x-y-z$ components of the ground reaction force assuming a point contact takes place between the robot and the ground substrate; $k_{1}$ and $k_{2}$ are the spring and damping coefficients of the compliant surface model; $\mu_c$, $\mu_s$, and $\mu_v$ are the Coulomb, static, and viscous friction coefficients; and, $v_s > 0$ is the Stribeck velocity. 

While active posture control of COBRA in simulation and hardware is currently possible, we keep the scope of this work limited to a fixed posture. However, in our next works, we plan to use this model to study the active steering of COBRA by manipulating its posture. One important step towards active steering is the identification of the contact point. The conjugate momentum $\sigma_{cm}$ obtained above will be employed to design estimators that can estimate contact location based on conjugate momentum observers. We leave further details for our next publications.   

In our simulations, we considered a joint position controller that fixates the robot's posture. Then, we simulated the dynamics of tumbling with an ODE solver in Matlab with the following integration details: The solver uses a variable step size with a maximum step size of $0.2\ seconds$. It has a relative tolerance of $1\times10^{-3}$ and an automatically set absolute tolerance.
The physical properties of the model simulation are as follows. Each module weighs $0.5\ kg$, including the head and tail modules, with each body module measuring $19.1\ cm$ in length, the head module measuring $17.5\ cm$ in length, and the tail module measuring $26.4\ cm$ in length. Each module has a diameter of $10\ cm$, except the tail module, which is designed to host an additional sensor payload and has a diameter of $14\ cm$. The diagonal components of the mass-moment of inertia, $I_{xx}$, $I_{yy}$, $I_{zz}$ ($kg.m^{2}$) are: the head ($9.2362\times10^{-4}$,$8.70128\times10^{-4}$,$5.79548\times10^{-4}$), tail ($22.3399\times10^{-4}$, $21.8196\times10^{-4}$, $13.1895\times10^{-4}$), and body modules ($8.51775\times10^{-4}$, $8.18543\times10^{-4}$, $5.89313\times10^{-4}$).


The ground model is flat and angled at a $10^\circ$ with a ground stiffness of $100 Nm$, a damping of $1\times 10^{-3} N/(m/s)$ and a transition region width of $1\times10^{-4} m$. 
Figure~\ref{fig:joint-torque-grf} shows the torques $\tau_i$ applied at the joints during the tumbling. The ground reaction forces $F_{GRF}$ are shown in Fig.~\ref{fig:joint-torque-grf}. As can be seen, the actuator torques corresponding to the contacting modules peaks offer insights into contact estimation models for future active control of tumbling. Figure~\ref{fig:force} shows the virtual forces needed to make a 1-deg-sideway tilt in COBRA during its tumbling. As this figure suggests, a relatively large axial load is needed to even make small changes in the plane of locomotion.



\section{Concluding Remarks}

Cobra's tumbling locomotion was put to a rigorous test at the NASA BIG Idea Forum in Pasadena, California. Early testing in readily available testing environments such as parking lot ramps showed partial success. At inclines of $4~deg.$, the robot could not build up enough speed to overcome irregularities of its own shape and would fall when rolling over cracks. On inclines of $8~deg.$, the robot was able to achieve higher speeds of tumbling and fell over less often on uneven ground. These were, however, short inclines of around $10~m$, and under passive tumbling, the rate of rotation was too slow to become stable. At the BIG Idea Forum, COBRA, along with other competing robots was given the opportunity to test at Lucerne Valley in California, where we had access to hilly and rocky terrain with a high degree of difficulty for mobile robot locomotion. The terrain was selected to best represent the goal of the BIG Idea competition, which was to demonstrate mobility in reaching the bottom of lunar craters on the south pole of the moon, which have steep uneven slopes with an average incline of $30~deg$. Under these conditions, COBRA successfully demonstrated tumbling locomotion on a rocky hill with an average incline of $24~deg$. Figure \ref{fig:exp} shows COBRA tumbling down the hill, beginning the locomotion by shifting its center of mass forward and going back to a hexagonal configuration once tumbling had been initiated. With the steep $24~deg.$ slope, the robot quickly picked up speed and became highly stable, making the full $50~m$ to the bottom of the hill in 10 seconds without tipping over.

\nocite{sihite_unsteady_2022, ramezani_generative_2021, sihite_unilateral_2021, lessieur_mechanical_2021, de_oliveira_thruster-assisted_2020, grizzle_progress_nodate, sihite_multi-modal_2023, sihite_orientation_2021, ramezani_towards_2020}
\printbibliography

@article{sihite_multi-modal_2023,
	title = {Multi-{Modal} {Mobility} {Morphobot} ({M4}) with appendage repurposing for locomotion plasticity enhancement},
	volume = {14},
	copyright = {2023 The Author(s)},
	issn = {2041-1723},
	url = {https://www.nature.com/articles/s41467-023-39018-y},
	doi = {10.1038/s41467-023-39018-y},
	language = {en},
	number = {1},
	urldate = {2023-11-18},
	journal = {Nature Communications},
	author = {Sihite, Eric and Kalantari, Arash and Nemovi, Reza and Ramezani, Alireza and Gharib, Morteza},
	month = jun,
	year = {2023},
	note = {Number: 1
Publisher: Nature Publishing Group},
	pages = {3323},
}

@inproceedings{lessieur_mechanical_2021,
	title = {Mechanical design and fabrication of a kinetic sculpture with application to bioinspired drone design},
	volume = {11758},
	url = {https://www.spiedigitallibrary.org/conference-proceedings-of-spie/11758/1175806/Mechanical-design-and-fabrication-of-a-kinetic-sculpture-with-application/10.1117/12.2587898.full},
	doi = {10.1117/12.2587898},
	urldate = {2023-05-17},
	booktitle = {Unmanned {Systems} {Technology} {XXIII}},
	publisher = {SPIE},
	author = {Lessieur, Andrew and Sihite, Eric and Dangol, Pravin and Singhal, Akshath and Ramezani, Alireza},
	month = apr,
	year = {2021},
	pages = {21--27},
}

@inproceedings{ramezani_generative_2021,
	title = {Generative {Design} of {NU}’s {Husky} {Carbon}, {A} {Morpho}-{Functional}, {Legged} {Robot}},
	doi = {10.1109/ICRA48506.2021.9561196},
	booktitle = {2021 {IEEE} {International} {Conference} on {Robotics} and {Automation} ({ICRA})},
	author = {Ramezani, Alireza and Dangol, Pravin and Sihite, Eric and Lessieur, Andrew and Kelly, Peter},
	month = may,
	year = {2021},
	note = {ISSN: 2577-087X},
	pages = {4040--4046},
}

@inproceedings{sihite_unilateral_2021,
	title = {Unilateral {Ground} {Contact} {Force} {Regulations} in {Thruster}-{Assisted} {Legged} {Locomotion}},
	doi = {10.1109/AIM46487.2021.9517648},
	booktitle = {2021 {IEEE}/{ASME} {International} {Conference} on {Advanced} {Intelligent} {Mechatronics} ({AIM})},
	author = {Sihite, Eric and Dangol, Pravin and Ramezani, Alireza},
	month = jul,
	year = {2021},
	note = {ISSN: 2159-6255},
	pages = {389--395},
}

@inproceedings{sihite_orientation_2021,
	title = {Orientation stabilization in a bioinspired bat-robot using integrated mechanical intelligence and control},
	volume = {11758},
	url = {https://www.spiedigitallibrary.org/conference-proceedings-of-spie/11758/1175805/Orientation-stabilization-in-a-bioinspired-bat-robot-using-integrated-mechanical/10.1117/12.2587894.full},
	doi = {10.1117/12.2587894},
	urldate = {2023-05-17},
	booktitle = {Unmanned {Systems} {Technology} {XXIII}},
	publisher = {SPIE},
	author = {Sihite, Eric and Lessieur, Andrew and Dangol, Pravin and Singhal, Akshath and Ramezani, Alireza},
	month = apr,
	year = {2021},
	pages = {12--20},
}

@inproceedings{sihite_unsteady_2022,
	title = {Unsteady aerodynamic modeling of {Aerobat} using lifting line theory and {Wagner}'s function},
	doi = {10.1109/IROS47612.2022.9982125},
	booktitle = {2022 {IEEE}/{RSJ} {International} {Conference} on {Intelligent} {Robots} and {Systems} ({IROS})},
	author = {Sihite, Eric and Ghanem, Paul and Salagame, Adarsh and Ramezani, Alireza},
	month = oct,
	year = {2022},
	note = {ISSN: 2153-0866},
	pages = {10493--10500},
}

@article{batten_kickbot_nodate,
	title = {Kickbot: {A} {Spherical} {Autonomous} {Robot}},
	language = {en},
	author = {Batten, Christopher and Wentzla, David},
}

@inproceedings{behar_nasajpl_2004,
	title = {{NASA}/{JPL} {Tumbleweed} polar rover},
	volume = {1},
	doi = {10.1109/AERO.2004.1367622},
	booktitle = {2004 {IEEE} {Aerospace} {Conference} {Proceedings} ({IEEE} {Cat}. {No}.{04TH8720})},
	author = {Behar, A. and Matthews, J. and Carsey, F. and Jones, J.},
	month = mar,
	year = {2004},
	note = {ISSN: 1095-323X},
	pages = {395 Vol.1},
}

@inproceedings{bicchi_introducing_1997,
	title = {Introducing the "{SPHERICLE}": an experimental testbed for research and teaching in nonholonomy},
	volume = {3},
	shorttitle = {Introducing the "{SPHERICLE}"},
	doi = {10.1109/ROBOT.1997.619356},
	booktitle = {Proceedings of {International} {Conference} on {Robotics} and {Automation}},
	author = {Bicchi, A. and Balluchi, A. and Prattichizzo, D. and Gorelli, A.},
	month = apr,
	year = {1997},
	pages = {2620--2625 vol.3},
}

@inproceedings{halme_motion_1996,
	title = {Motion control of a spherical mobile robot},
	volume = {1},
	doi = {10.1109/AMC.1996.509415},
	booktitle = {Proceedings of 4th {IEEE} {International} {Workshop} on {Advanced} {Motion} {Control} - {AMC} '96 - {MIE}},
	author = {Halme, A. and Schonberg, T. and Wang, Yan},
	month = mar,
	year = {1996},
	pages = {259--264 vol.1},
}

@article{javadi_a_introducing_2004,
	title = {Introducing {Glory}: {A} {Novel} {Strategy} for an {Omnidirectional} {Spherical} {Rolling} {Robot}},
	volume = {126},
	issn = {0022-0434},
	shorttitle = {Introducing {Glory}},
	url = {https://doi.org/10.1115/1.1789542},
	doi = {10.1115/1.1789542},
	number = {3},
	urldate = {2023-02-21},
	journal = {Journal of Dynamic Systems, Measurement, and Control},
	author = {Javadi A., Amir Homayoun and Mojabi, Puyan},
	month = dec,
	year = {2004},
	pages = {678--683},
}

@inproceedings{mukherjee_simple_1999,
	title = {Simple motion planning strategies for spherobot: a spherical mobile robot},
	volume = {3},
	shorttitle = {Simple motion planning strategies for spherobot},
	doi = {10.1109/CDC.1999.831235},
	booktitle = {Proceedings of the 38th {IEEE} {Conference} on {Decision} and {Control} ({Cat}. {No}.{99CH36304})},
	author = {Mukherjee, R. and Minor, M.A. and Pukrushpan, J.T.},
	month = dec,
	year = {1999},
	note = {ISSN: 0191-2216},
	pages = {2132--2137 vol.3},
}

@article{paskarbeit_ourobotsensorized_2021,
	title = {Ourobot—{A} sensorized closed-kinematic-chain robot for shape-adaptive rolling in rough terrain},
	volume = {140},
	issn = {0921-8890},
	url = {https://www.sciencedirect.com/science/article/pii/S0921889020305558},
	doi = {10.1016/j.robot.2020.103715},
	language = {en},
	urldate = {2023-02-23},
	journal = {Robotics and Autonomous Systems},
	author = {Paskarbeit, Jan and Beyer, Simon and Engel, Matthäus and Gucze, Adrian and Schröder, Johann and Schneider, Axel},
	month = jun,
	year = {2021},
	pages = {103715},
}

@inproceedings{reina_rough-terrain_2004,
	title = {Rough-terrain traversability for a cylindrical shaped mobile robot},
	doi = {10.1109/ICMECH.2004.1364428},
	booktitle = {Proceedings of the {IEEE} {International} {Conference} on {Mechatronics}, 2004. {ICM} '04.},
	author = {Reina, G. and Foglia, M. and Milella, A. and Gentile, A.},
	month = jun,
	year = {2004},
	pages = {148--153},
}

@article{sastra_dynamic_2009,
	title = {Dynamic {Rolling} for a {Modular} {Loop} {Robot}},
	volume = {28},
	issn = {0278-3649},
	url = {https://doi.org/10.1177/0278364908099463},
	doi = {10.1177/0278364908099463},
	number = {6},
	urldate = {2023-02-26},
	journal = {The International Journal of Robotics Research},
	author = {Sastra, Jimmy and Chitta, Sachin and Yim, Mark},
	month = jun,
	year = {2009},
	note = {Publisher: SAGE Publications Ltd STM},
	pages = {758--773},
}

@article{sugiyama_crawling_2006,
	title = {Crawling and {Jumping} by a {Deformable} {Robot}},
	volume = {25},
	issn = {0278-3649},
	url = {https://doi.org/10.1177/0278364906065386},
	doi = {10.1177/0278364906065386},
	number = {5-6},
	urldate = {2023-02-23},
	journal = {The International Journal of Robotics Research},
	author = {Sugiyama, Yuuta and Hirai, Shinichi},
	month = may,
	year = {2006},
	note = {Publisher: SAGE Publications Ltd STM},
	pages = {603--620},
}

@article{tian_dynamic_2015,
	title = {Dynamic rolling analysis of triangular-bipyramid robot},
	volume = {33},
	issn = {0263-5747, 1469-8668},
	url = {https://www.cambridge.org/core/journals/robotica/article/dynamic-rolling-analysis-of-triangularbipyramid-robot/F74BD0B65670FE9BF50BD742DB5DB7DE},
	doi = {10.1017/S0263574714000666},
	language = {en},
	number = {4},
	urldate = {2023-02-21},
	journal = {Robotica},
	author = {Tian, Yaobin and Yao, Yan-An},
	month = may,
	year = {2015},
	note = {Publisher: Cambridge University Press},
	pages = {884--897},
}

@article{wang_dynamics_2018,
	title = {Dynamics of a rolling robot of closed five-arc-shaped-bar linkage},
	volume = {121},
	issn = {0094-114X},
	url = {https://www.sciencedirect.com/science/article/pii/S0094114X17302379},
	doi = {10.1016/j.mechmachtheory.2017.10.010},
	language = {en},
	urldate = {2023-02-21},
	journal = {Mechanism and Machine Theory},
	author = {Wang, Yujin and Wu, Changlin and Yu, Lianqing and Mei, Yuanyuan},
	month = mar,
	year = {2018},
	pages = {75--91},
}

@article{wang_trajectory_2018,
	title = {Trajectory planning of a rolling robot of closed five-bow-shaped-bar linkage},
	volume = {53},
	issn = {0736-5845},
	url = {https://www.sciencedirect.com/science/article/pii/S0736584517300340},
	doi = {10.1016/j.rcim.2018.03.004},
	language = {en},
	urldate = {2023-02-21},
	journal = {Robotics and Computer-Integrated Manufacturing},
	author = {Wang, Yujin and Wu, Changlin and Yu, Lianqing and Mei, Yuanyuan},
	month = oct,
	year = {2018},
	pages = {81--92},
}

@article{wei_design_2019,
	title = {Design and locomotion analysis of a novel modular rolling robot},
	volume = {133},
	issn = {0094-114X},
	url = {https://www.sciencedirect.com/science/article/pii/S0094114X18306281},
	doi = {10.1016/j.mechmachtheory.2018.11.004},
	language = {en},
	urldate = {2023-02-21},
	journal = {Mechanism and Machine Theory},
	author = {Wei, Xiangzhi and Tian, Yaobin and Wen, Shanshan},
	month = mar,
	year = {2019},
	pages = {23--43},
}

@article{western_golden_2023,
	title = {Golden wheel spider-inspired rolling robots for planetary exploration},
	volume = {204},
	issn = {0094-5765},
	url = {https://www.sciencedirect.com/science/article/pii/S0094576522007020},
	doi = {10.1016/j.actaastro.2022.12.030},
	language = {en},
	urldate = {2023-02-23},
	journal = {Acta Astronautica},
	author = {Western, A. and Haghshenas-Jaryani, M. and Hassanalian, M.},
	month = mar,
	year = {2023},
	pages = {34--48},
}

@misc{noauthor_big_nodate,
	title = {Big {Idea} {\textbar} {NASA}'s {Breakthrough}, {Innovative}, and {Game}-changing ({BIG}) {Idea} {Challenge}},
	url = {https://bigidea.nianet.org/},
	language = {en-US},
}

@inproceedings{de_oliveira_thruster-assisted_2020,
	title = {Thruster-assisted {Center} {Manifold} {Shaping} in {Bipedal} {Legged} {Locomotion}},
	doi = {10.1109/AIM43001.2020.9158967},
	booktitle = {2020 {IEEE}/{ASME} {International} {Conference} on {Advanced} {Intelligent} {Mechatronics} ({AIM})},
	author = {de Oliveira, Arthur C. B. and Ramezani, Alireza},
	month = jul,
	year = {2020},
	note = {ISSN: 2159-6255},
	pages = {508--513},
}

@inproceedings{ramezani_towards_2020,
	title = {Towards biomimicry of a bat-style perching maneuver on structures: the manipulation of inertial dynamics},
	shorttitle = {Towards biomimicry of a bat-style perching maneuver on structures},
	doi = {10.1109/ICRA40945.2020.9197376},
	booktitle = {2020 {IEEE} {International} {Conference} on {Robotics} and {Automation} ({ICRA})},
	author = {Ramezani, Alireza},
	month = may,
	year = {2020},
	note = {ISSN: 2577-087X},
	pages = {7015--7021},
}

@article{grizzle_progress_nodate,
	title = {Progress on {Controlling} {MARLO}, an {ATRIAS}-series {3D} {Underactuated} {Bipedal} {Robot}},
	language = {en},
	author = {Grizzle, J W and Ramezani, A and Buss, B and Griﬃn, B and Hamed, K Akbari and Galloway, K S},
}

\end{document}